# Using CODEQ to Train Feed-forward Neural Networks

Mahamed G. H. Omran[1] and Faisal al-Adwani[2]

[1] Department of Computer Science, Gulf University for Science and Technology, Kuwait, Kuwait
`omran.m@gust.edu.kw`
[2] Petroleum Engineering Department, Kuwait University, Kuwait, Kuwait
`f.adwani@ku.edu.kw`

**Abstract.** CODEQ is a new, population-based meta-heuristic algorithm that is a hybrid of concepts from chaotic search, opposition-based learning, differential evolution and quantum mechanics. CODEQ has successfully been used to solve different types of problems (e.g. constrained, integer-programming, engineering) with excellent results. In this paper, CODEQ is used to train feed-forward neural networks. The proposed method is compared with particle swarm optimization and differential evolution algorithms on three data sets with encouraging results.

## 1 Introduction

Artificial neural networks (ANNs) provide a way to learn arbitrary mapping between two data sets. A neural network contains a number of adjustable parameters called weights and biases. Data sets consisting of input patterns and their corresponding output values are used to train the network. The objective of training a neural network is to minimize the mapping error that is the difference between the output value as specified in the data set and the output of the network given the corresponding input pattern.

Both local and global optimization methods have been used to train ANNs. The most popular training methods are variants of gradient based back-propagation algorithms (e.g. modified back-propagation (Liang *et al*. 1994), back propagation via conjugate gradient (Charalambous 1992), etc), which are local methods. Local search algorithms depend on initial conditions and are prone to be trapped in a local optimum.

To address the limitations of local search methods, global search approaches have been used to train ANNs. Global optimizers such as particle swarm optimization (PSO) (Kennedy and Eberhart 1995) and differential evolution (DE) (Storn and Price 1995) have been successfully applied to the problem of training ANNs (Gudise and Venayagamoorthy 2003; Masters and Land 1997; Ilonen *et al*. 2007).

Motivated by the advantages of PSO, DE and other population-based methods, a recently-proposed, parameter-free populaton-based algorithm, called CODEQ, is used to train feed-forward ANNs. CODEQ (Omran and Salman 2009) is a metaheuristic approach that is built based on of concepts from chaotic search, opposition-based learning (Tizhoosh 2005), DE and quantum mechanics. The performance of CODEQ has already been investigated and compared with other well-known population-based optimization approaches (e.g. DE and PSO) when applied to eleven benchmark functions (Omran 2009). The results show that CODEQ provides excellent results with the added advantage of no parameter tuning. The application of CODEQ to constrained problems was investigated by Omran and Salman (2009) with encouraging results. Furthermore, CODEQ was successfully used to solve integer programming problems (Omran and al-Sharhan 2009).

The reminder of the paper is organized as follows: Section 2 provides an overview of CODEQ. The proposed method is presented in Section 3. Data sets to measure the performance of the different approaches are provided in Section 4. Results of the experiments are presented in Section 5. Finally, Section 6 concludes the paper.

## 2 CODEQ

The CODEQ algorithm (Omran and Salman 2009) works as follows:

*Step 1*. A population of *s* vectors are randomly initialized within the search space.

*Step 2*. For each parent, $x_i(t)$, of iteration *t*, a trial vector, $v_i(t)$, is created by mutating the parent vector. Two individuals $x_{i_1}(t)$, and $x_{i_2}(t)$ are randomly selected with $i_1 \neq i_2 \neq i$, and the difference vector, $x_{i_1} - x_{i_2}$, is calculated. The trial vector is then calculated as

$$v_i(t) = x_i(t) + (x_{i_1}(t) - x_{i_2}(t))\ln\left(\frac{1}{u}\right) \quad (1)$$

where $u \sim U(0,1)$.

The generated offspring, $v_i(t)$, replaces the parent, $x_i(t)$, only if the fitness of the offspring is better than that of the parent (i.e. apply greedy selection).

*Step 3*. For each iteration *t*, a new vector is created as,

$$w(t) = \begin{cases} LB + UB - r \times x_b(t) & \text{if } U(0,1) \leq 0.5 \\ x_g(t) + |x_{i_1}(t) - x_{i_2}(t)| \times (2c(t) - 1) & \text{otherwise} \end{cases} \quad (2)$$

where $r \sim U(0,1)$, *LB* and *UB* are the lower and upper bounds of the problem, $x_b(t)$ is the worst (i.e. least fit) vector in iteration *t*, $x_g(t)$ is the best (i.e. fittest) vector in iteration *t*, $x_{i_1}(t)$, and $x_{i_2}(t)$ are randomly selected vectors with $i_1 \neq i_2 \neq i$ and $c(t)$ is a chaotic variable defined as,

$$c(t) = \begin{cases} c(t-1)/p & c(t-1) \in (0,p) \\ (1-c(t-1))/(1-p) & c(t-1) \in [p,1) \end{cases}$$

where $c(0)$ and $p$ are initialized randomly within the interval (0,1).

*Step 4*. The generated vector, $w_i(t)$, replaces the worst vector in iteration $t$, $x_b(t)$, only if the fitness of $w_i(t)$ is better than that of $x_b(t)$.

*Step 5*. Repeat steps 2-4 until a stopping criterion is satisfied.

## 3  The proposed method

CODEQ is a floating-point encoded optimization algorithm for global optimization over continuous spaces. Thus, it can be used within the weight and bias spaces of a feed-forward neural network.

In network training, each vector consists of $N$ network weights and $B$ biases:

$$x_i(t) = (w_{1,i,t}, \ldots, w_{N,i,t}, b_{1,i,t}, \ldots, b_{B,i,t}), \quad i = 1, \ldots, s$$

The objective (i.e. fitness) function that needs to be minimized is the training set mean squared error (MSE) of the feed-forward neural network. The MSE is calculated as follows,

$$MSE = \sum_{i=1} (\hat{X}_i - X_i)^2$$

where
$\hat{X}_i$: The estimated values from ANN
$X_i$: The actual value

## 4  Data sets

The proposed method is tested against three different data sets where the ANN model parameters are optimized in order to find the best solution. In addition, each data set has been divided into a training set and a testing set. The training data set is used to train the network and optimize weights and biases. The testing data set is used to test the optimized parameters. The data sets are as follow:

**4.1 House**

This data set estimate the median value of owner occupied homes in Boston suburbs given 13 neighborhoods attributes. There are 506 samples which have been di-vided into 430 samples used to train the ANN model and the 76 samples used for validation.

### 4.2 Oil Price

The oil price data set consists of the West Texas crude spot price in terms of different parameters which include crude oil production, crude oil supply, crude oil demand and refinery capacities and throughput. The data used in this study starts from January-1980 to December-2003. In order to capture the effect of the combined effects the model was built with creating indices for each factor normalized to production. That is for each demand or supply level it has been normalized with average monthly production. The data set consists of 288 samples in which 244 samples user for training and the remaining used to validate the model.

### 4.3 IRIS

The IRIS data set classify iris flowers based on four attributes with an output of three different flower types. The data set consists of 150 samples in which 127 used to training and the remaining 73 used to validate the model.

## 5 Experimental Results

This section compares the performance of CODEQ with that of particle swarm optimization (PSO) and self-adaptive differential evolution (SDE) (Brest *et al*. 2006). SDE is a new version of DE where the control parameters are self-adaptive (i.e. SDE requires no parameter tuning). Results show that SDE generally outperformed DE and other evolutionary algorithms from literature when applied to 21 benchmark functions (Brest *et al*. 2006).

The results reported in this section are averages and standard deviations over 30 simulations. Each simulation was allowed to run for 10,000 evaluations of the objective function (i.e. MSE) using a population size of 20 individuals (i.e. $s = 20$). In order to make a fair comparison, the same initial population was used for all algorithms. The statistically significant best solutions have been shown in bold (using the non-parametric statistical test called Wilcoxon's rank sum test for independent samples (Wilcoxon 1945) with $\alpha = 0.05$).

Table 1 shows results for all data sets which include training and testing data sets. As shown in the table it is clearly shown that the CODEQ algorithm has the best estimation of the set of weights and biases that result in the minimum MSE values as shown in table 1 for both training data set and testing data set. For example, for data set #1, the testing data set gives the best MSE value of 33.11 when using the CODEQ algorithm compared to 341.94 for PSO and 74.38 for SDE algorithms. It should be noted that it is better to check the testing data set since those data has not been used to optimize the weight and biases values hence they are used for prediction purposes.

In addition Figure 1 shows summary of network for training data sets for all data sets. It is clearly shown that the CODEQ algorithm outperform both PSO and SDE algorithm where it yields the minimum MSE values. Figure 2 shows MSE values and again it is clearly shown that the CODEQ algorithm will better enhance network performance than both PSO and SDE algorithms

Table 1: Summary of MSE Results for the Three Algorithms

| Data Type | Data Set | PSO | SDE | CODEQ |
|---|---|---|---|---|
| Training | 1 | 48.61(12.41) | 71.06(14.22) | **33.06(1.147)** |
|  | 2 | 33.14(14.52) | 49.40(11.01) | **26.88(13.31)** |
|  | 3 | 0.15(0.06) | 4.44(8.28) | **0.05(0.02)** |
| Testing | 1 | 341.94(1414.95) | 74.38(15.196) | **33.11(12.70)** |
|  | 2 | 4637.14(13554.16) | 5283.84(28689.73) | **435.25(1544.31)** |
|  | 3 | 9.60(29.42) | 20.74(44.04) | **0.03(0.02)** |

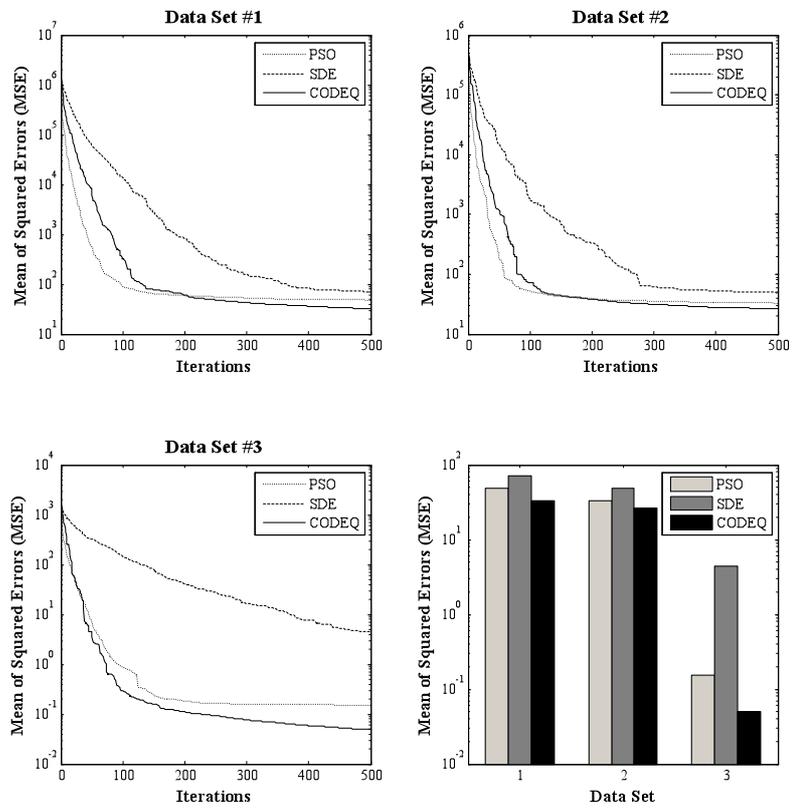

**Figure 1: ANN Preformance for Training Data Sets**

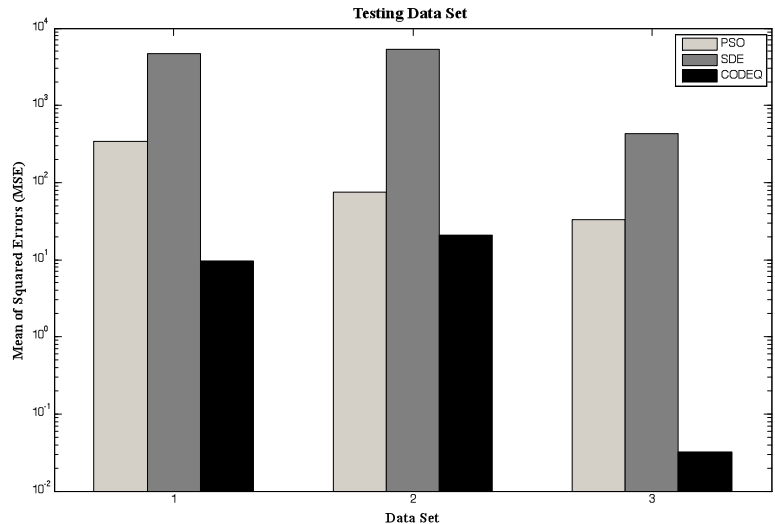

**Figure 2: Model Performance for Testing Data Set**

## 5  Conclusion

In this paper, CODEQ was used to train feed-forward neural networks. The proposed method was compared with PSO and DE on three data sets. The results showed that CODEQ outperformed the other approaches.

Future work will study the effect of using a local search algorithm within CODEQ to train ANNs. According to our preliminary results using the back-propagation training method to refine the best vector in each iteration of CODEQ yields very promising results.